\documentclass[10pt,twocolumn,letterpaper]{article}

\usepackage{wacv}
\usepackage{times}
\usepackage{epsfig}
\usepackage{graphicx}
\usepackage{amsmath}
\usepackage{amssymb}
\usepackage{booktabs}

\newcommand{\argmin}{\mathop{\rm arg~min}\limits}

\newcommand{\sinha}[1]{{\color{black}#1}}

\def\wacvPaperID{320} 

\wacvfinalcopy 

\ifwacvfinal
\usepackage[breaklinks=true,bookmarks=false]{hyperref}
\else
\usepackage[pagebackref=true,breaklinks=true,colorlinks,bookmarks=false]{hyperref}
\fi

\begin{document}

\title{Difficulty-Net: Learning to Predict Difficulty for Long-Tailed Recognition}

\author{Saptarshi Sinha\\
Hitachi Ltd.\\
Tokyo, Japan\\
{\tt\small saptarshi.sinha.hx@hitachi.com}
\and
Hiroki Ohashi\\
Hitachi Ltd.\\
Tokyo, Japan\\
{\tt\small hiroki.ohashi.uo@hitachi.com}
}

\maketitle
\thispagestyle{empty}

\begin{abstract}
   Long-tailed datasets, where head classes comprise much more training samples than tail classes, cause recognition models to get biased towards the head classes. 
   Weighted loss is one of the most popular ways of mitigating this issue, and a recent work has suggested that class-difficulty might be a better clue than conventionally used class-frequency to decide the distribution of weights.
   A heuristic formulation was used in the previous work for quantifying the difficulty, but we empirically find that the optimal formulation varies depending on the characteristics of datasets.
   Therefore, we propose Difficulty-Net, which learns to predict the difficulty of classes using the model's performance in a meta-learning framework.
   To make it learn reasonable difficulty of a class within the context of other classes, we newly introduce two key concepts, namely the relative difficulty and the driver loss. 
   The former helps Difficulty-Net take other classes into account when calculating difficulty of a class, while the latter is indispensable for guiding the learning to a meaningful direction.
   Extensive experiments on popular long-tailed datasets demonstrated the effectiveness of the proposed method, and it achieved state-of-the-art performance on multiple long-tailed datasets.
\end{abstract}

\section{Introduction}\label{sec:intro}

Despite the outstanding performance of the recent deep learning (DL) models on public datasets, deploying such models in the real world often leads to a performance drop. 
One of the causes is that the public datasets are usually almost perfectly class-balanced while real-world data are generally long-tailed, where a few classes (called \textit{head classes}) consist of a significantly larger number of training samples than the rest of the classes (called \textit{tail classes}).
The `long-tailed recognition' research domain particularly aims at addressing this issue.

\begin{figure}[t]
  \centering
    \includegraphics[width=\linewidth]{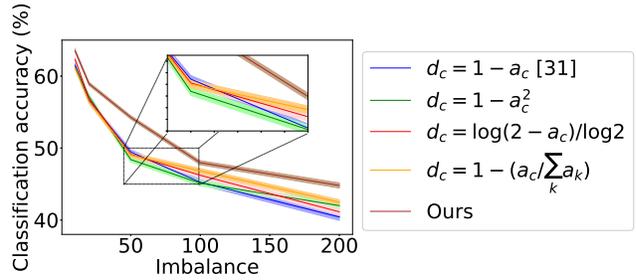}
  \caption{Different quantifications of class-difficulties perform better than others in different situations (imbalance ratios). The imbalance in the data is  calculated as the ratio of the frequency of the most frequent class to that of the least frequent class.
 We compute class-wise difficulty ($d_c$) using four different functions of class-wise accuracy ($a_c$) for the CDB-CE~\cite{cdb-ce} loss function and compare their performance on the CIFAR100-LT. 
 Interestingly, 
the alternate formulations work better than the originally proposed one ($d_c = 1 - a_c$) in many cases.
However, the best performing function changes with the imbalance values. This brings us to the question ``Which formulation to choose for my imbalanced dataset?" 
}
  \label{fig:class-wise difficulty}
\end{figure}

Amongst multiple possible strategies to tackle long-tailed recognition problems, cost-sensitive learning is one of the most popular and promising strategies.
Most cost-sensitive learning techniques modify the cost function to penalize the model differently for different samples. 
This modification is generally done by scaling the cost value using different weights, and the research direction is mainly aimed at finding an effective weight-assignment strategy. 
One simple and intuitive way is to assign weights using the inverse of the class-frequencies. 
Recently, more sophisticated approaches such as class-balanced loss~\cite{classbalancedloss} and equalization loss~\cite{eqlloss} have been proposed.
However, most of these approaches give more weights to the tail classes because they assume that tail classes are always the most difficult to learn. 
\sinha{Recently Sinha \etal~\cite{cdb-ce,sinha_ijcv}} empirically showed that the above assumption does not always hold true and further claimed that class-difficulty might be a better clue to decide weights. 

While they proposed an intuitive quantification of the class-level difficulties, 
this quantification is preliminarily determined regardless of the property of a given dataset, and thus may not be optimal in different situations.
In fact, we empirically found that multiple quantifications for class-wise difficulty gave comparable or even better results than~\cite{cdb-ce} as shown in Figure~\ref{fig:class-wise difficulty}. 
This adds the extra tedious task of selecting the appropriate formulation for a given imbalanced data. 

Motivated by recently proposed Meta-Weight-Net (MWN)~\cite{meta-weight-net}, our research aims to address the above issue by meta-learning a simple model, named Difficulty-Net, to predict class-level difficulty scores and then dynamically distribute the weights based on the scores.
Such a strategy removes dependence on any prior formulation for class-wise difficulty and lets the model learn any suitable function to compute it.
The key difference with MWN is three folds.
First, while MWN is a sample-level weighting method, ours is a class-level weighted approach, whose advantage in long-tailed recognition has been revealed in~\cite{cdb-ce} and also discussed in Sec.~\ref{sec: diff from mwn and cdb-ce} and Sec.~\ref{sec:ablation}. 
Second, we propose to use relative difficulties rather than absolute difficulties that are used in prior works~\cite{meta-weight-net,cdb-ce} so that the other classes' difficulties are also taken into account when determining the difficulty of a class.
Third, we propose a new loss function that drives the learning process of Difficulty-Net in a reasonable direction, without which the performance turned out to degrade.



To summarize, our key contributions are:
\vspace{-5pt}
\begin{itemize}
\itemsep0em
    \item We propose Difficulty-Net, which learns to predict class-difficulty in a meta-learning framework.
    \item We argue that relative difficulty is more important and effective than absolute difficulty, and provide an empirical evidence for the argument.
    \item We propose a new loss function, called driver loss, that guides the learning process in a reasonable direction.
    \item We conducted extensive experiments on multiple long-tail benchmark datasets and achieved state-of-the-art results. In addition, we provide in-depth analysis on the effect and property of the proposed method in comparison to previous works, which revealed the effectiveness of our method.
\end{itemize}


\section{Related works}\label{sec:related works}

Major strategies to tackle the long-tailed recognition can be broadly categorized as 
\sinha{data re-sampling methods~\cite{smote,borderline-smote,classbalancedsampling_1,oversampling_1,undersampling_1,sinha_ijcv}}, 
metric learning~\cite{tripletloss,contrastivelearning,liftedloss,m-sloss},
knowledge transfer~\cite{oltr,kt_1,kt_2}, 
mixture of experts~\cite{RIDE,LFME,TADE}, 
\sinha{cost-sensitive learning~\cite{ldam-drw,classbalancedloss,logitadjustment,meta-weight-net,sinha_ijcv}} and decoupled learning~\cite{decoupling,dro-lt,disalign,mislas}.

Data re-sampling techniques~\cite{smote,borderline-smote,classbalancedsampling_1,oversampling_1,undersampling_1} try to neutralize the long-tail by under-sampling from head classes or over-sampling from tail classes. 
Under-sampling~\cite{undersampling_1} generally results in poor representation of the head classes,
while a straight-forward over-sampling strategy of replicating tail-class samples causes the model to overfit on the repeated samples.
Another popular oversampling technique is synthetic data generation~\cite{smote,borderline-smote} for the tail classes. 
Certain class-balanced sampling approaches such as class-aware sampling~\cite{classbalancedsampling_1,decoupling,classbalancedsampling_2} and square-root sampling~\cite{decoupling} have been shown to be more effective than using over- or under-sampling.
They typically try to increase the sampling rate for the tail classes during training. 
However, they still result in overfitting due to the repeated sampling of the same samples from tail classes.

Metric learning methods~\cite{hingeloss,tripletloss,contrastivelearning,liftedloss,m-sloss,rangeloss} aim to learn a high-quality feature extractor that preserves inter-class and intra-class relationships in the feature space. 
They achieve this by learning from pairs~\cite{contrastivelearning,liftedloss,rangeloss} or triplets~\cite{hingeloss,tripletloss} of input samples. 
Metric learning has been used in long-tailed recognition~\cite{oltr,hybridnet} in hope that high-quality feature extractor will mitigate the imbalance between head and tail classes. 
An effective sampling of the sample groups is the key for efficient training in this scheme. 
However, such sampling strategies come with the risk of under-representation or overfitting, as explained above.

Knowledge transfer~\cite{featuretransfer2,oltr,kt_1,kt_2,featuretransfer} in long-tailed recognition tries to transfer knowledge gained from the head classes to the tail classes. They achieve this either by learning modular transformations from few-shot model parameters to many-shot models~\cite{kt_1,kt_2} or by designing external modules for feature transfer~\cite{oltr,featuretransfer}.
Designing such modules is usually computationally expensive in real-world usecases~\cite{decoupling}.

Mixture of experts (MoE)~\cite{RIDE,LFME,TADE,bbn} is an ensemble-based technique where the expert models are trained to gain diverse knowledge. The aggregated knowledge of the experts is either used directly to alleviate the long-tail~\cite{RIDE,TADE} or used to teach a student model for that purpose~\cite{LFME}. 
Despite the increasing popularity of this domain, our research focuses on the improvement of a single 
model 
as
it can then easily be combined with any mixture.

Cost-sensitive learning can be achieved by logit-adjustment loss~\cite{ldam-drw,logitadjustment,seesawloss} and weighted loss~\cite{classbalancedloss,hardmining,focalloss,l2rw,meta-weight-net,eqllossv2,eqlloss} approaches. 
Most prior methods distribute these adjustment or weight values on the basis of class-frequencies. 
Recently, \sinha{Sinha \etal~\cite{cdb-ce,sinha_ijcv}} showed that class-difficulty is a better metric for the purpose. 
However, finding an optimal formulation for calculating class-difficulty is not a trivial task as the optimal formulation usually varies depending on datasets as shown in Figure~\ref{fig:class-wise difficulty}.
Our research builds on the work of Sinha \etal~\cite{cdb-ce} and tries to remove the requirement of any prior formulation by using meta-learning. Meta-learning~\cite{metalearning_1} has previously been used in long-tailed recognition to learn~\cite{jamal,l2rw} or predict~\cite{meta-weight-net} sample weights. The closest to our research is Meta-Weight-Net (MWN)~\cite{meta-weight-net}, which learns a model to predict sample-level weights from training loss. 
Different from them, we use class-level weighting, which is known to be better than sample-level weighting in long-tailed recognition~\cite{cdb-ce}.

Recently, Kang \etal~\cite{decoupling} found that decoupling model learning into representation learning and classifier learning helps long-tailed recognition. Since this finding, most works have tried to improve either the data-representation~\cite{dro-lt,ssl} or the classifier~\cite{groupsoftmax,classifierbalancing,disalign,mislas}. 
We show that our method also benefits from this framework and achieves state-of-the-art results based on it.

\section{Proposed method}
\subsection{Background}
Generally, the prior assumption is that the tail classes are the most difficult to learn for the models. However, it has recently been empirically shown that the number of training instances of a class might not be the best clue to determine its difficulty because some classes are well-represented even with fewer training samples. On the basis of this finding, Sinha \etal~\cite{cdb-ce} came up with a simple formulation to directly calculate the difficulty of a class from the model's performance. The formulation says that if the model's classification accuracy on a class $c$ is $a_c$, then the difficulty of the class $d_c$ can be computed as $d_c = 1 - a_c$.

However, we found two lacking points in 
the formulation. 
First, as stated in Sec.~\ref{sec:intro}, we found multiple decreasing functions of accuracy $a_c$ that outperformed the above formulation in multiple setups.
In the meantime, we also found that the best performing formulation varies inconsistently with the data imbalance and thus it is not possible to preliminarily define the best formulation for a given dataset. 
Second, while the above formulation helps to compute the absolute difficulty of a class, we believe it is more important to compute the difficulty of a class {\it relative} to the other classes because it is reasonable to assign a high difficulty score to a class with high accuracy (\ie easy class) {\it if} the other classes have even higher accuracies. 
For that purpose, all the classes need to be considered when computing the difficulty of a single class, which is not done in~\cite{cdb-ce}. 

To address these issues, we propose meta-learning the formulation that is most effective for a given dataset, taking relative difficulties into consideration.


\subsection{Meta-learning via Difficulty-Net}\label{sec:meta-learning_via_difficulty-net}
\paragraph{Difficulty-Net design.}\label{difficultynet design}
Given a dataset of $C$ classes, we aim to learn a formulation that can compute the relative difficulty for each class. For this purpose, we design the formulation for class-wise difficulty as 
\begin{equation}
    d_1, d_2 , \dots, d_C = D(\{a_c\}_{c=1}^C; \theta).
    \label{difficultynet}
\end{equation}
Note that both $d_c$ and $a_c$ change as the training progresses, but here we omit the notation of training steps for simplicity.
$D$ is a neural network with parameters $\theta$. 
In our implementation, we choose $D$ to be a simple MLP model with two hidden layers.
The output layer dimension is kept same as the number of classes and
a sigmoid activation at the output ensures the difficulty scores to be in the range $(0,1)$.
The input to $D$ are the model's classification accuracies for all $C$ classes. The design of $D$ ensures that while estimating the difficulty for a class, the model's performance on the other classes is also taken into account. We refer to $D$ as `Difficulty-Net'.


\paragraph{Meta-learning objective:}
Suppose a classification problem in which we are provided a training dataset $S^{train} = \{x_i, y_i\}_{i=1}^N$, where $x_i$ is the $i^{th}$ training sample and $y_i \in \{1,\dots, C\}$ is its corresponding ground truth label. Given a classifier neural network $f(x; \phi)$ with learnable parameters $\phi$, our primary objective is to learn the optimal parameters $\phi^*$ so that $f(x;\phi^*)$ provides the minimum classification loss on the training set $S^{train}$, \ie 
\begin{equation}
    \phi^* = \argmin_{\phi} \frac{1}{N} \sum_{i=1}^N L( f(x_i; \phi), y_i),
    \label{typical optimization}
\end{equation}
where $L$ computes the loss corresponding to $f$'s prediction for a given sample and is typically the cross-entropy loss.

In long-tailed recognition, the training dataset $S^{train}$ is class-imbalanced.
In such cases, optimization using Eq.~\ref{typical optimization} leads to biased learning of $\phi$. To compensate for the imbalance, we modify the learning objective as most weighted loss approaches do, \ie
\begin{equation}
    \phi^* = \argmin_{\phi} \frac{1}{N} \sum_{i=1}^N w_i L( f(x_i; \phi), y_i),
    \label{weighted loss optimization}
\end{equation}
where $w_i$ is the weight assigned to the training sample $x_i$. 
In our proposed approach, $w_i$ is computed by 
Difficulty-Net
$D$ as 
\begin{equation}
    w_i(A_C, \theta) = D(A_C; \theta)_{y_i},
    \label{weight computation}
\end{equation}
where $D(A_C; \theta)_{y_i}$ is the difficulty score for class $y_i$ predicted by Difficulty-Net and $A_C = \{a_c\}_{c=1}^C$ is the set of accuracies of $f(x; \phi)$ for all $C$ classes, evaluated prior to this calculation. 
Therefore, the learning objective for $\phi^*$ is modified as 
\begin{equation}
    \phi^*(\theta) = \argmin_{\phi} \frac{1}{N} \sum_{i=1}^N w_i(A_C, \theta) L( f(x_i; \phi), y_i).
    \label{our loss optimization}
\end{equation}

Since the optimization of our main classifier network depends on the effectiveness of Difficulty-Net, it is important to optimize the parameters $\theta$ of $D$ as well. 
Inspired by~\cite{meta-weight-net}, we use a small balanced meta-dataset $S^{meta} = \{x^{meta}_i, y^{meta}_i\}_{i=1}^M$ for optimizing the parameters $\theta$ as follows.
\begin{multline}
        \theta^* = \argmin_{\theta}\frac{1}{M} \sum_{i=1}^M  L^{meta}_i(\phi^*(\theta)) \\=\argmin_{\theta} \frac{1}{M} \sum_{i=1}^M  L( f(x^{meta}_i; \phi^* (\theta)), y^{meta}_i).
    \label{meta loss optimization}
\end{multline}

However, we found that $L^{meta}$ alone is not enough for Difficulty-Net to learn to estimate difficulties from accuracy. 
Even for 2 classes with very different accuracy values, Difficulty-Net learned using Eq.~\ref{meta loss optimization} tends to give similar difficulty scores for both classes.
To address this issue, we add another loss component to drive the learning of Difficulty-Net in a practically correct direction. We call this loss `driver loss' and calculate it as
\begin{equation}
    L^{dr} (A_C,\theta) = \frac{1}{C}\sum_{c=1}^C \left(\left(1 - \hat{a_c}\right) -  D(A_C; \theta)_{c}\right)^2,
    \label{driver loss}
\end{equation}
where $\hat{a_c} = {a_c}/{\sum_k a_k}$ is the normalized accuracy of class $c$.
The $L^{dr}$ is built on the motivation that Difficulty-Net should learn to give high difficulty scores to a class, if the accuracy of the class is relatively low.
Now the parameters $\theta$ of Difficulty-Net $D$ are optimized as 
\begin{equation}
    \theta^* = \argmin_{\theta} \lambda L^{dr}(A_C, \theta) + \frac{1}{M} \sum_{i=1}^M  L^{meta}_i(\phi^*(\theta)), \label{eq:lambda}
\end{equation}
where $\lambda$ is a hyper-parameter controlling the influence of $L^{dr}$. Note that too high value of $\lambda$ will simply cause $D$ to always predict class difficulties as 
$d_c = 1 - \hat{a_c}$.
We ablate over various values of $\lambda$ in our experiments.

\paragraph{Learning method.}

Following~\cite{meta-weight-net}, our meta-learning method is a 3-step process. Given a classifier network $f(x; \phi_t)$ and Difficulty-Net $D(; \theta_t)$ at time step $t$, the first step aims to learn 
intermediate classifier parameter $\hat{\phi_t}$ by
\begin{equation}
    \hat{\phi_t}(\theta_t) \xleftarrow[]{}  \phi_t -  \alpha \frac{1}{b} \sum_{i=1}^b w_i(A_{C,t}, \theta_t) \frac{\partial L( f(x_i; \phi), y_i)}{\partial\phi}\Big|_{\phi_t},
    \label{step 1}
\end{equation}
where $\alpha$ is the step size for gradient descent and $b$ is the number of samples in one mini-batch sampled from the training set $S^{train}$. 
$A_{C,t}$ is the classification accuracy of $f(x; \phi_t)$ on all the $C$ classes at time step $t$ and is computed on a validation dataset $S^{val}$. 

The second step updates the parameters $\theta$ of Difficulty-Net using the obtained intermediate classifier $f(x; \hat{\phi_t})$ on a mini-batch of size $m$ sampled from the meta-dataset. The update is done by
\begin{multline}
       \theta_{t+1} \xleftarrow[]{} \\\theta_t - 
         \hspace{-0.04in} \beta \frac{\partial (\lambda L^{dr}(A_{C,t}, \theta) + \frac{1}{m} \sum\limits_{i=1}^m  L^{meta}_i(\hat{\phi_t}(\theta)))}{\partial \theta} \Big|_{\theta_t},
        \label{step 2}
        \end{multline}
where $\beta$ is the step size for updating the parameters of Difficulty-Net.

Finally, the third step uses the updated parameters $\theta_{t+1}$ to update the parameters of the classifier network $f(x; \phi_t)$ over the same mini-batch sampled in Eq.~\ref{step 1}. 
\begin{equation}
    \phi_{t+1} \xleftarrow{} \phi_t - \alpha\frac{1 }{b} \sum_{i=1}^b w_i(A_{C,t}, \theta_{t+1}) \frac{\partial L( f(x_i; \phi), y_i)}{\partial\phi}\Big|_{\phi_t}.
    \label{step 3}
\end{equation}
The above three steps are executed iteratively till convergence or the end of the training. 
The overall algorithm is presented in 
Algorithm~1
in the supplementary material.

For our experiments, we construct the $S^{meta}$ following exactly the same procedure as~\cite{jamal,meta-weight-net}. We also found that $S^{meta}$ is reusable as $S^{val}$ for calculating $A_{C,t}$. Therefore, we do not use any extra data compared to previous methods.
Also, although it is ideal to calculate $A_{C,t}$ for every time step $t$, we calculate the accuracy only after every epoch in our implementation for saving computational time.
\paragraph{Difference with MWN and CDB-CE.}\label{sec: diff from mwn and cdb-ce}
Although we share a similar meta-learning framework as MWN~\cite{meta-weight-net}, our approach is very different from theirs in more than one way.
One difference is that MWN is a sample-level weighting strategy while ours is a class-level weighting strategy. 
The advantages of class-level weighting over sample-level weighting in long-tailed learning is pointed out in~\cite{cdb-ce} and also reflected in our experimental results. 

Ours is not the straight-forward combination of MWN~\cite{meta-weight-net} and CDB-CE~\cite{cdb-ce}.
First, both MWN and CDB-CE use absolute difficulties of a sample or a class to determine the weights. 
We believe, however, the relative difficulty compared with other samples or classes is more important because it is reasonable to assign a high difficulty score to a class with high accuracy (\ie easy class) {\it if} the other classes have even higher accuracies. 
The proposed method estimates relative difficulties of each class amongst all the classes, and it turned out to be more effective as we will show in Sec.~\ref{sec:ablation}.
Second, the straight-forward combination of these prior works without the driver loss turns out to learn almost nothing and predicts almost identical difficulties for all the classes as we will show in Sec.~\ref{sec:ablation}. 
The newly proposed driver loss is essential to guide the training in a reasonable direction.

The empirical evidences of these arguments are provided in Sec.~\ref{sec:ablation} and~\ref{sec:further_analysis}.

\section{Experiments}
\subsection{Datasets}
     
\paragraph{\bf CIFAR100-LT.} 
CIFAR100~\cite{CIFAR} is an object-centric balanced classification data-set comprised of tiny images belonging to 100 different classes. Long-tailed versions of the dataset are artificially created by reducing the training samples per class according to an exponential function as given in~\cite{classbalancedloss}. Following~\cite{classbalancedloss}, we use CIFAR100-LT with imbalance varying in 10--200. 
\vspace{-11pt}
\paragraph{\bf ImageNet-LT.} 
ImageNet-LT is a long-tail version of ImageNet~\cite{imagenet} created by~\cite{oltr}. It contains 1000 object categories with heavy imbalance of 256.
We use the same \textit{train, val} and \textit{ test} splits as~\cite{oltr}. 
\vspace{-11pt}
\paragraph{\bf Places-LT.} 
Places-2~\cite{places} is a large-scale scene-centric image dataset,
used for scene recognition tasks. Places-LT is a long-tailed subset of Places-2 with 365 classes and imbalance of 996, created by~\cite{oltr}. 
We use the same splits as~\cite{oltr}. 

For constructing $S^{meta}$, we followed the setup of previous meta-learning based methods \cite{jamal,meta-weight-net} to ensure the fair comparison. Please see the supplementary material for the details.
The evaluation results are reported on balanced test sets.

\subsection{Implementation details}
Following previous long-tailed works~\cite{classbalancedloss,oltr,eqlloss}, we use ResNet-32~\cite{ResNet} for  CIFAR-100-LT experiments. 
On ImageNet-LT, we follow~\cite{decoupling,oltr,mislas} and use ResNet-10~\cite{ResNet}, ResNet-50~\cite{ResNet}. 
As in~\cite{mislas}, we use pretrained (on ImageNet~\cite{imagenet}) ResNet-152~\cite{ResNet} and finetune it on Places-LT. 
The basic architecture of Difficulty-Net is the same for all the datasets as explained in Sec.~\ref{difficultynet design}, \ie MLP with two hidden layers, but we change the dimension of hidden and output layers as different datasets have different number of classes.
We will explain a simple way to select the dimension of hidden layers in the supplementary material. 
We evaluate our method both in end-to-end (e2e) learning and decoupled learning~\cite{decoupling} settings.
For using Difficulty-Net in decoupled learning, we first train the respective model using Difficulty-Net based weighting. 
Then, following~\cite{decoupling}, we freeze the feature extractor and re-train the classifier without using Difficulty-Net.
We use $\lambda=0.3$ for all the experiments unless otherwise stated since we find it works reasonably well as we will show in Sec.~\ref{sec:ablation}. Further details are provided in the supplementary material.

\subsection{Compared methods}
For comparison, we use multiple SOTA methods including (1) data-resampling: class-balanced sampling (CB sampling)~\cite{decoupling}, (2) cost-sensitive learning: equalization loss (EQL)~\cite{eqlloss}, focal loss~\cite{focalloss}, class-balanced loss~\cite{classbalancedloss}, label-distribution-aware-margin (LDAM) loss~\cite{ldam-drw}, pre-formulated class-difficulty balanced loss (CDB-CE)~\cite{cdb-ce}, (3) metric learning: parametric contrastive learning (PaCo)~\cite{PaCo}, (4) decoupled learning: classifier normalization ($\tau$-norm)~\cite{decoupling}, classifier re-training (cRT)~\cite{decoupling}, learnable weight scaling (LWS)~\cite{decoupling}, label-aware smoothing (LAS)~\cite{mislas}, balanced meta-softmax (BALMS)~\cite{BALMS}, distribution robustness loss (DRO-LT)~\cite{dro-lt}, (5) meta learning: Meta-Weight-Net (MWN)~\cite{meta-weight-net}, class-balancing as domain-adaptation (CB-DA)~\cite{jamal}. 
For the sake of fairness, we do not compare our method directly with MoE methods~\cite{RIDE,LFME} as they use ensemble of multiple expert models, while we focus on improving the learning for a single expert. 
However, we verified that the proposed method can exhibit significant performance gains by using simple ensembling techniques and can 
outperform SOTA MoE methods. The results are found in the supplementary
material.

\subsection{Main results}
\label{sec:main_results}
\paragraph{\bf CIFAR100-LT.} 

\begin{table}[t]
  \begin{center}
    {\small{
\begin{tabular}{llllll}
\toprule
&\multicolumn{5}{c}{Imbalance}\\
    \midrule
    Method &  200 & 100 & 50 & 20 & 10\\
    \midrule
    \textit{e2e training} \\
    Focal Loss~\cite{focalloss} & 39.64 & 44.03 & 48.91 & 55.57 & 61.10  \\
    MWN~\cite{meta-weight-net} & 40.25 & 44.81 & 49.68 & 56.53 & 61.44\\
    Class-Balanced~\cite{classbalancedloss} &39.95 & 44.78 & 47.67 & 56.83 & 59.95\\
    CB-DA~\cite{jamal} & 40.89& 46.24 & 49.80 & 56.67 & 62.16\\
    LDAM~\cite{ldam-drw} & 41.42 & 46.14 & 49.19 & 55.90 & 62.08\\
    EQL~\cite{eqlloss} & 43.46 & 46.47 & 51.34 & 56.82 & 60.13\\
    CDB-CE~\cite{cdb-ce} & 40.42 & 45.25 & 49.45 & 56.66 & 61.52 \\
    PaCo~\cite{PaCo} & 43.09 & 47.26 & 52.14 & 58.37 & 63.12\\
    \hspace{1mm} + Bal.~Softmax~\cite{BALMS} & 46.72 & 51.47 & 55.88 & 60.32 & 64.10 \\
    \hspace{1mm} + Bal.~Softmax~\cite{BALMS}$\dagger$ & -- & 52.00 & 56.00 & -- & 64.20\\
    Ours & 44.80 & 47.96 & 54.27 & 58.93 & 63.52 \\
    \hspace{1mm} + Bal.~Softmax & 47.53 & 52.14 & 56.86 & \textbf{61.72} & \textbf{65.67} \\
    \midrule
    \textit{decoupled learning}\\
     cRT~\cite{decoupling} & 44.05 & 48.04 & 53.32 & 58.72 & 63.74 \\
     LWS~\cite{decoupling} & 44.42 & 48.13 & 53.44 & 59.10 & 63.97\\
     LAS~\cite{mislas} & 44.87& 48.68 & 53.85&59.36 & 64.18\\
     DRO-LT~\cite{dro-lt} $\dagger$ &-- & 47.31 & \textbf{57.57} & -- & 63.41 \\
     BALMS~\cite{BALMS}  & 46.12 & 50.95 & 54.42 & 59.00 & 63.10\\
     MWN + cRT &44.56 & 48.34 &53.62 & 59.05 & 63.99 \\
     MWN + LWS &44.71& 48.65 & 53.77&59.22 & 64.15 \\
     MWN + LAS & 45.04& 49.12 &53.95 &59.38 & 64.24\\
     Ours + cRT & 47.45 & 52.01 & 56.34 & 61.08& 64.80\\ 
     Ours + LWS & \underline{47.91} & \underline{52.62} & 56.61 & 61.38 & 65.08\\ 
     Ours + LAS & \textbf{48.32} &\textbf{52.96} & \underline{56.90} & \underline{61.46} & \underline{65.22}\\
\bottomrule
\end{tabular}
}}
\end{center}
\caption{Top-1 classification accuracy (\%) on CIFAR-100-LT. 
  $\dagger$ denotes copied results from origin paper~\cite{PaCo,dro-lt}. The best results are made bold while the second best results are underlined, which applies for the other tables as well.}
  \label{tab:CIFAR-LT results}
\end{table}
Following~\cite{PaCo,eqlloss}, we use AutoAugment~\cite{AutoAugment} and Cutout~\cite{Cutout} for all our implementations on CIFAR100-LT. As explained in~\cite{eqlloss}, this achieves a higher baseline than other commonly followed ones. Therefore, to ensure the fairness of comparison, we re-implemented the compared methods in our training setup using their published codes. 
Results without using AutoAugment and Cutout are provided in the supplementary material.
We achieved better results than originally reported results for all the re-implemented methods except PaCo~\cite{PaCo}, which uses additional augmentation. 
Therefore, we list the original results of PaCo for reference in addition to the results in the fair setting.
PaCo uses an additional center learning rebalance step, for which they employ Balanced Softmax (Bal.~Softmax)~\cite{BALMS}. 
We report the results of PaCo both with and without the use of Bal.~Softmax.
For the fair comparison with PaCo + Bal.~Softmax, we tested Ours + Bal.~Softmax in addition to the vanilla variant (Ours). 

In e2e learning, our proposed approach without combining any other techniques (Ours) achieved better performance than all previous stand-alone methods as seen in Table~\ref{tab:CIFAR-LT results}. 
The margin of improvement is higher in high-imbalanced situations.
End-to-end learning with Ours + Bal.~Softmax turned out to be very effective and
created new SOTA for low imbalanced cases (\ie 10 and 20). 

In decoupled-learning, we find that when we use feature extractors trained using Difficulty-Net, any popular classifier learning method (\eg cRT, LWS, LAS) gives improved performance. 
This shows that our proposed method learns very powerful data representations. 
Ours + LAS achieved the best results in high imbalanced situations (\eg 200 and 100), while it achieved the second best in all other cases.

\paragraph{\bf ImageNet-LT.} 
Table~\ref{tab:ImageNet-LT results} shows the results on ImageNet-LT.
In e2e learning alone, irrespective of the model used, we achieved better overall accuracy than other e2e 
methods and comparable accuracy with multiple decoupled 
methods.

Furthermore, Difficulty-Net based representation learning with popular classifier re-training methods achieved state-of-the-art results. Using both ResNet-10 and ResNet-50, Ours + LAS achieved the best overall accuracy, which re-confirms the effectiveness of this method. More ImageNet-LT results with many-/med-/few-shot splits are available in the supplementary material.

\begin{table}
  \begin{center}
    {\small{
\begin{tabular}{llr}
\toprule
         Method & ResNet-10 & ResNet-50 \\
         \midrule
         \textit{e2e training} \\
         CE  & 34.8 & 41.6 \\
         Focal loss~\cite{focalloss} & 30.5 & -- \\
         EQL~\cite{eqlloss}  & 36.4 & --\\
         CB-DA\cite{jamal} & 36.7 & 48.0\\
         CDB-CE~\cite{cdb-ce} & 38.5 & --\\
          Bal.~Softmax~\cite{BALMS} & 41.1 & --\\
         PaCo~\cite{PaCo} $^*$ & -- &  49.8\\
         \hspace{0.2mm} + Bal.~Softmax~\cite{BALMS}$^*$ & -- & 53.5\\ 
         Ours & 41.4 & 51.2 \\
         \hspace{0.2mm} + Bal.~Softmax & 44.3 & \underline{53.7} \\
         \hline
         \textit{decoupled learning}\\
         
         cRT~\cite{decoupling}  & 41.8 & 47.3 \\
         LWS~\cite{decoupling}  & 41.4 & 47.7 \\
         MiSLAS~\cite{mislas}  & -- & 52.7 \\
         BALMS~\cite{BALMS}  & 41.8 & -- \\
         DRO-LT~\cite{dro-lt}  & -- & 53.5 \\
         Ours + cRT & 43.6 & 53.5 \\
         Ours + LWS & \underline{44.4} & \underline{53.7} \\
         Ours + LAS & \textbf{44.6} & \textbf{54.0}\\
         \bottomrule

\end{tabular}
}}
\end{center}
\caption{Top-1 classification accuracies (\%) on ImageNet-LT. $^*$ represents reproduced results using author's codes without using RandAugment~\cite{randaugment} for fair comparison. Other baseline results are copied from original papers. Results using RandAugment
are provided in the supplementary
material.
}
  \label{tab:ImageNet-LT results}
\end{table}

\paragraph{\bf Places-LT.} 
From Table~\ref{tab:CIFAR-LT results} and Table~\ref{tab:ImageNet-LT results}, it is evident that our Difficulty-Net based weighting is consistently effective when used for the representation learning in decoupled training methods. 
Therefore, for Places-LT, we only report the results of Ours + \{cRT, LWS, LAS\} and compare them with previous SOTA results in Table~\ref{tab:Places results}. 
The results verify that the representation learned using our method is very powerful and helps us achieve the best overall accuracy by simple classifier re-balancing. Our improvements in overall accuracy is majorly accounted for by significant gains in medium- and few-shot accuracies. 
Even though our representation learning is effective with any classifier re-training method, especially Ours + LAS significantly boosts results for the few-shot classes and achieved SOTA in overall accuracy. PaCo achieved the best results for the medium-shot classes, but it sacrificed the performance on the many-shot classes significantly, resulting in lower overall accuracy. 
\begin{table}
  \begin{center}
    {\small{
\begin{tabular}{llllr}
\toprule
    Method &  Many & Med & Few & All\\
    \midrule
    CE & \underline{45.7} & 27.3 & 8.2 & 30.2 \\
    CB sampling~\cite{decoupling} & -- & -- & -- & 30.3\\
    Focal Loss~\cite{focalloss} & 41.1 & 34.8 & 22.4 & 34.6 \\
     cRT~\cite{decoupling} & 42.0 & 37.6  & 24.9 & 36.7 \\
     LWS~\cite{decoupling} & 40.6 & 39.1 & 28.6 & 37.6 \\
     BALMS~\cite{BALMS} & 41.2 & 39.8 & 31.6 & 38.7\\
     LADE~\cite{lade} & 42.8& 39.0& 31.2& 38.8\\
     DisAlign~\cite{disalign} & 40.4 & 42.4 & 30.1 & 39.3 \\
     IEM~\cite{IEM} &\textbf{46.8}& 39.2& 28.0 & 39.7\\
     MiSLAS~\cite{mislas} & 39.6 & 43.3 & 36.1 & 40.4\\
     PaCo~\cite{PaCo} & 37.5 & \textbf{47.2} & 33.9 & 41.2 \\
     
    
     Ours + cRT &43.0 & \underline{43.8} & 35.0& \textbf{41.7}\\ 
     Ours + LWS & 41.4 & 43.7 & \textbf{36.9} & \underline{41.5}\\ 
     Ours + LAS & 42.4 & 43.7 & \underline{36.6} & \textbf{41.7} \\
    \bottomrule

\end{tabular}
}}
\end{center}
\caption{Top-1 classification accuracies (\%) for Places-LT. }
  \label{tab:Places results}
\end{table}

\subsection{Ablation study}\label{sec:ablation}
In this sub-section, we first show the ablation study of our key components, namely relative difficulty and the driver loss.
Then we re-verify the effectiveness of the class-level weighting studied in~\cite{cdb-ce} in our meta-learning framework. Further, we verify the effectiveness of using the meta-learning loss in our method.
We conclude this sub-section with the effectiveness of the proposed method by comparing it with the straight-forward combination of CDB-CE~\cite{cdb-ce} and MWN~\cite{meta-weight-net}.

\begin{table*}[]
  \begin{center}
    {\small{
\begin{tabular}{lllllllr}
\toprule
  \# & Name & S vs.~C & A vs.~R & ML & $L^{dr}$ & Imb.=100 & Imb.=10\\
  \midrule
  1 & Focal loss~\cite{focalloss}   & S & A &               &  & 44.03 & 61.10\\
  2 & CDB-CE~\cite{cdb-ce}          & C & A &               &  & 45.25 & 61.52 \\
  3 & MWN~\cite{meta-weight-net}    & S & A & \checkmark    &  & 44.81 & 61.44 \\
  4 & CDB-CE + MWN                  & C & A & \checkmark    &  & 45.42 & 61.87\\
  5 & Ours w/o $L^{dr}$ and ML  & C & R &               &  & 45.51 & 62.44\\
  6 & Ours w/o $L^{dr}$         & C & R & \checkmark    &  & 46.40 & \underline{63.10}\\
  7 & Ours w/o relative difficulty  & C & A & \checkmark    &  \checkmark & \underline{46.81} & 62.32\\
  8 & Ours w/o class-level weighting& S & R & \checkmark & \checkmark & 45.76 & 62.51\\
  9 & Ours                          & C & R & \checkmark & \checkmark & \textbf{47.96} & \textbf{63.52}\\
  \bottomrule
\end{tabular}
}}
\end{center}
\caption{Classification accuracy on CIFAR100-LT with imbalance ratio (Imb.) 100 and 10. ``S vs.~C'' means Sample-level vs.~Class-level. ``A vs.~R'' means Absolute difficulty vs.~Relative difficulty. ML stands for Meta Learning. }
  \label{tab:ablation}
\end{table*}

\paragraph{\bf Absolute difficulty vs.~relative difficulty.} 
For predicting absolute difficulty, we modified Difficulty-Net from Eq.~\ref{difficultynet} to $d_c = D^{abs}(a_c; \theta)$ and trained it in the same way as before. 
The comparison is provided in Table~\ref{tab:ablation} (\#7 vs.~\#9).
As can be seen, relative difficulty significantly outperforms absolute difficulty for both low and high imbalance. 
This verifies the effectiveness of relative difficulty.


\paragraph{\bf Contribution of $L^{dr}$.} 
The value of $\lambda$ in Eq.~\ref{eq:lambda} controls the impact of $L^{dr}$. 
Here we analyse the effect of $\lambda$.
For that, we evaluate the performance of ResNet-32 trained end-to-end using different values of $\lambda$ and report the results in Table~\ref{tab:analysis for diff imbalance}.
\begin{table}
  \begin{center}
    {\small{
    \begin{tabular}{lllllr}
\toprule
    $\lambda$ & 0 & 0.3  & 0.6  & 0.9  & 1.0 \\
    \midrule
    Imbalance=100 & 46.40 & \underline{47.96} & \textbf{48.03} & 47.35 & 46.66\\
    Imbalance=10 & 63.10 & \textbf{63.52} & \underline{63.44} & 62.62 & 62.24\\
     \bottomrule

\end{tabular}
}}
\end{center}
\caption{Accuracy (in e2e learning) for different values of $\lambda$ on CIFAR100-LT.  }
  \label{tab:analysis for diff imbalance}
\end{table}
It shows that $\lambda=0$, which is equivalent to \#6 in Table~\ref{tab:ablation}, works significantly poor especially in high imbalance case, which asserts the importance of using $L^{dr}$. We find that for higher imbalance, higher $\lambda$ works better. 
But, too high $\lambda$ leads to significant drop in performance.
Irrespective of  the imbalance, $\lambda=0.3$ works consistently well.

\begin{figure}[t]
\begin{center}
    \includegraphics[width=\linewidth ]{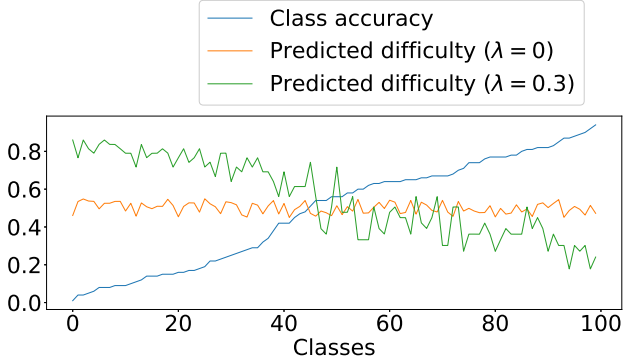}
\end{center}
\caption{Difficulty scores for CIFAR100-LT (imbalance=100) classes predicted by Difficulty-Net learned with $\lambda = 0$ and $\lambda=0.3$. The classes are sorted in increasing order of their accuracy. }
\label{fig:difficulty-net}
\end{figure}

 To further analyse the usefulness of $L^{dr}$, we visualise the predicted difficulty by Difficulty-Net trained with and without $L^{dr}$. The results are shown in Figure~\ref{fig:difficulty-net}. 
It shows that using $L^{dr}$ with $\lambda=0.3$ provides a more meaningful learning of Difficulty-Net compared to when not using $L^{dr}$ ($\lambda = 0$).
The latter predicts similar difficulty scores for all the classes inspite of the highly biased accuracy. 
However, using $L^{dr}$ helps Difficulty-Net to predict high difficulty for less accurate classes. 

\vspace{-8pt}
\paragraph{\bf Sample-level difficulty vs.~Class-level difficulty.}
We modified the Difficulty-Net to predict sample-level difficulties and compared it with the proposed method.
We modified the Difficulty-Net as $d_1, d_2, ..., d_B = D^{sample}(\{l_s\}_{s=1}^B; \theta)$ where $B$ is the total number of samples in a single batch, $l_s$ is the model's cross-entropy loss for samples $s$, and $d_s$ is the predicted difficulty for sample $s$. 
Simply, we meta-learn the $D^{sample}$ to predict difficulty of each sample relative to other samples in the same training batch.
Note that this variant uses relative difficulty and the driver loss, and thus is different from MWN.

In Table~\ref{tab:ablation} (\#8 vs.~\#9), it is seen that class-level difficulty significantly outperforms the sample-level difficulty in overall performance. This proves the effectiveness of class-difficulty in our proposed method. 
We believe that this happens because the head classes have higher {\it absolute} number of hard samples than the tail classes simply because the head classes have much more training samples. 
In such case, as pointed out in~\cite{cdb-ce}, sample-level weighting gives higher weights to head classes in total, and therefore cause the model to get biased to the head classes. 
This is verified by the fact that class-level performs much better especially for the tail classes (med and few-shot) as shown in the supplementary material. 
Another interesting observation is that our sample-level Difficulty-Net even significantly outperforms MWN (\#3 vs.~\#8), which re-verifies the effectiveness of our newly proposed components, namely relative difficulty and the driver loss.
\paragraph{\bf Contribution of $L^{meta}$.}
Here we verify the usefulness of $L^{meta}$ in our Difficulty-Net training. In Table~\ref{tab:ablation} (\#5 vs. \#6), we see that using meta-learning loss gives a boost of 0.89\% (for imb. 100) which confirms the benefit of using ML.
\paragraph{\bf The straight-forward combination of CDB-CE and MWN does not work.}
As stated in Sec.~\ref{sec: diff from mwn and cdb-ce}, ours is not the straight-forward combination of the previous methods. Evidently from Table~\ref{tab:ablation} (\#4 vs.~\#9), such straight-forward combination does not work well, which verifies the contributions of our newly proposed components.

\subsection{Further analysis}\label{sec:further_analysis}

\begin{figure}[t]
\begin{center}
    \includegraphics[width=\linewidth]{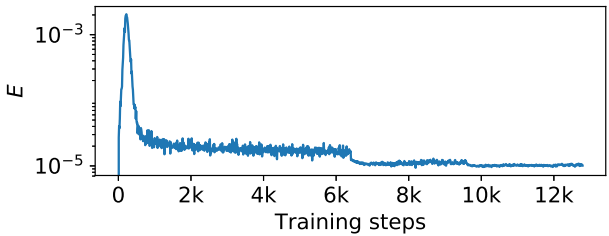}
\end{center}
\caption{Plotting entropy ($E$) of difficulty scores predicted by Difficulty-Net against number of training steps. We used CIFAR100-LT (imbalance=100) for this plot. }
\label{fig:weight_entropy}
\end{figure}

It is evident in Figure~\ref{fig:difficulty-net} that Difficulty-Net successfully learns to predict reasonable difficulty from the class-wise accuracies.
Here we further analyse how the predicted difficulties change as the training progresses.
For this purpose, we plot the entropy of the difficulty scores with the training steps in Figure~\ref{fig:weight_entropy}. 
We compute the entropy as $E(\{d_c\}_{c=1}^C) = -\frac{1}{C}\sum_{c=1}^C \log(C\frac{d_c}{\sum_k d_k})$. 
Figure~\ref{fig:weight_entropy} shows that the entropy 
decreases with the training steps. 
This suggests that the predicted difficulty scores gradually become more and more uniform, as the model's class-wise performance 
gradually gets balanced.
this result empirically supports the rationality of Difficulty-Net.

\begin{figure}[t]
  \begin{center}

  \begin{tabular}{cc}
     \includegraphics[width=0.45\linewidth]{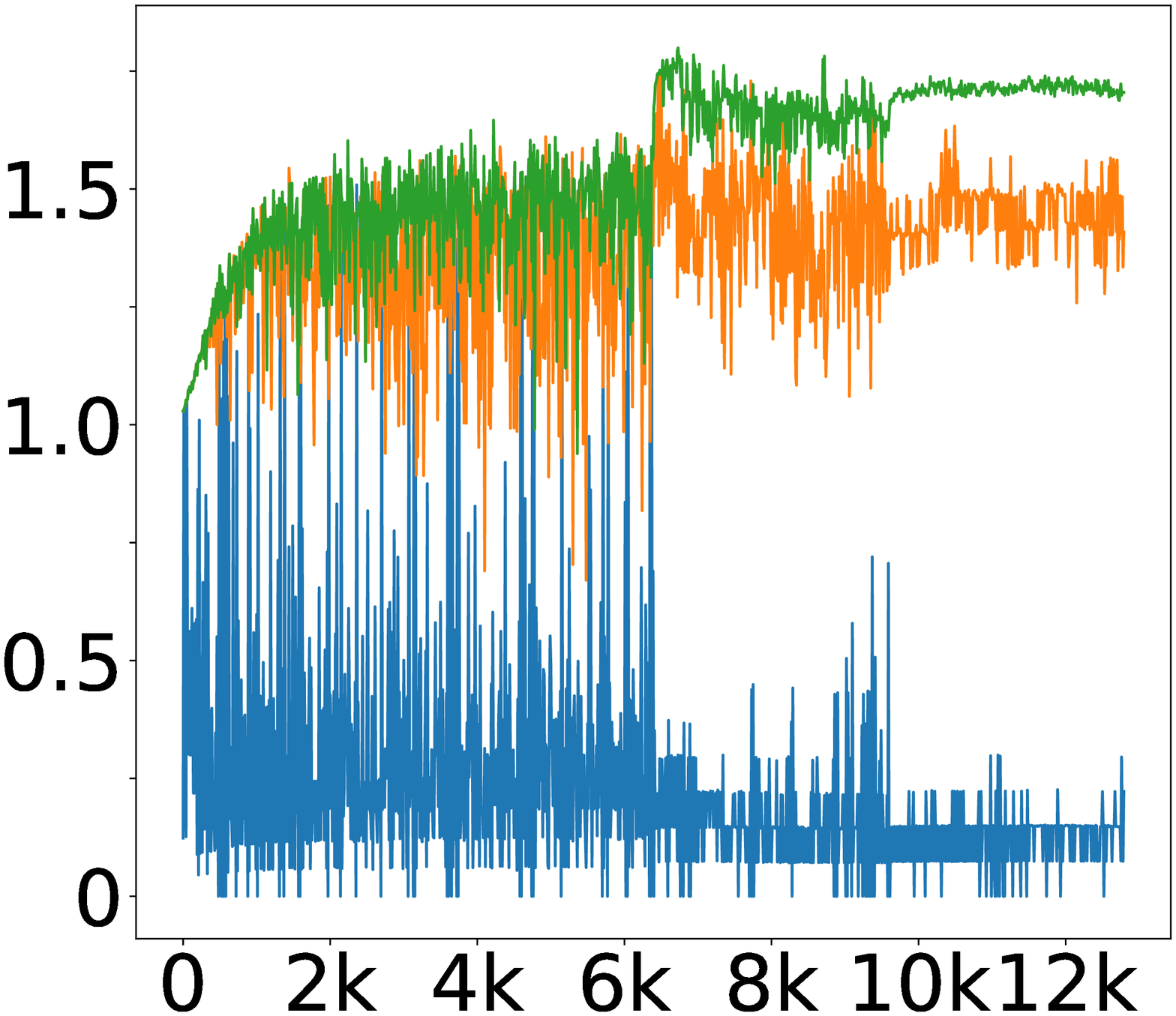}  
  \hfill
    \includegraphics[width=0.45\linewidth,  ]{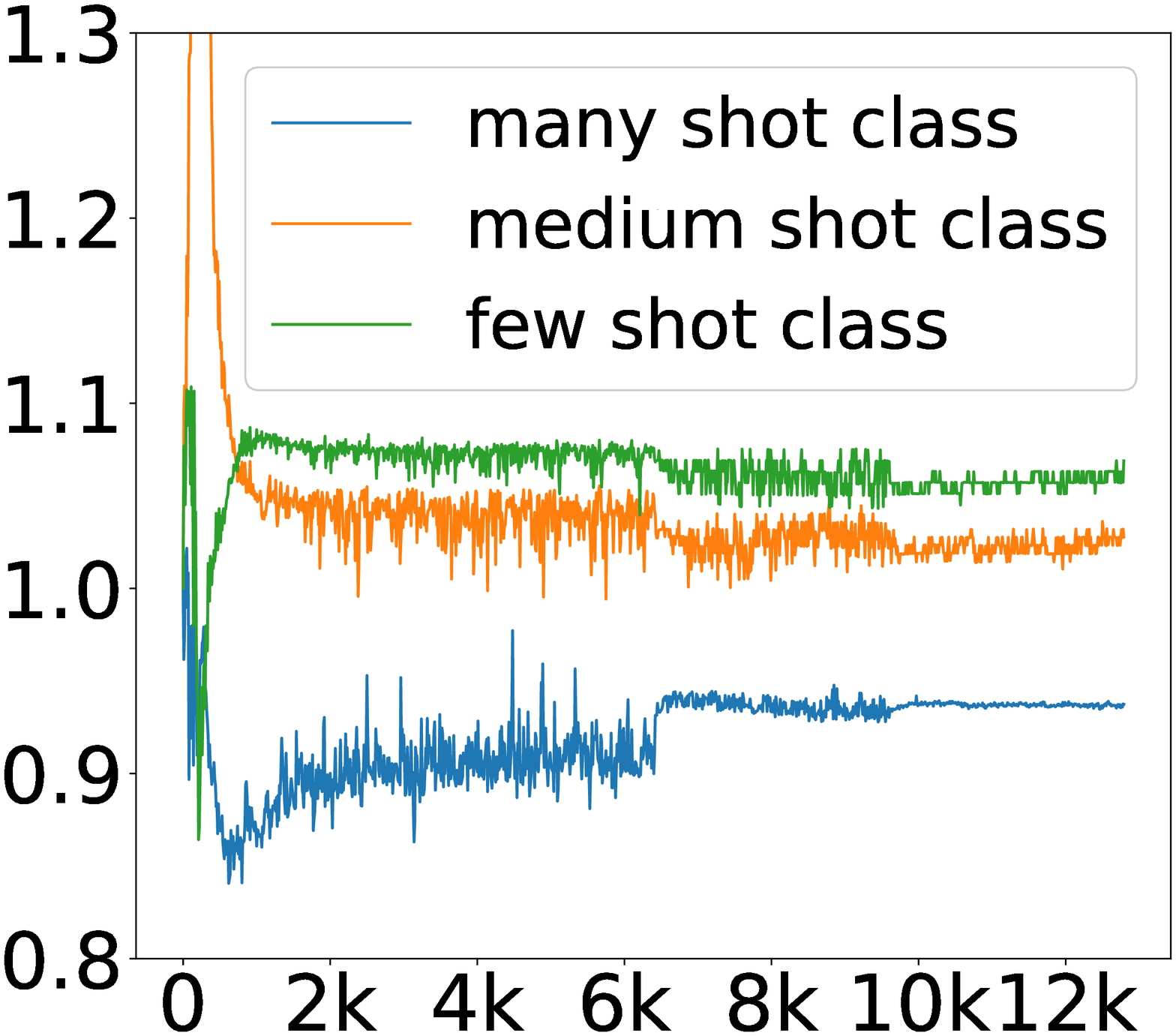} 
     \end{tabular}
  \caption{Assigned weights to three different classes during training with CDB-CE~\cite{cdb-ce} (Left) and our Difficulty-Net (Right). The vertical axis represents assigned weights and the horizontal axis represents training steps.}
  \label{weight-analysis}
  \end{center}
\end{figure}

Further, we analyse the characteristics of difficulties estimated by Difficulty-Net in comparison with those by~\cite{cdb-ce}. 
We pick three classes from the CIFAR100-LT classes (one from each of many-, medium- and few-shot classes) and show how the normalized weights of the classes change as the training progresses.
As shown in Figure~\ref{weight-analysis}, CDB-CE weighting~\cite{cdb-ce} leads to more fluctuations in the assigned weights, while Difficulty-Net based weighting is more smooth and stable. This suggests that Difficulty-Net has capability of `remembering' which class is difficult whereas CDB-CE weighting tends to be heavily affected by quick accuracy change at each time step. We believe this characteristic of Difficulty-Net encourages consistent and stable training of the model, ending up in better performance than CDB-CE weighting. Another interesting observation in Figure~\ref{weight-analysis} is that the difficulties of the three classes estimated by Difficulty-Net tend to converge as the training progresses 
,which is not observed in the case of CDB-CE. 
This observation is consistent with Figure~\ref{fig:weight_entropy}, which showed that the predicted difficulty scores gradually become more uniform as the 
model's performance gets balanced.

\section{Conclusion}

This paper has proposed Difficulty-Net, a novel method for long-tailed recognition that learns to predict difficulty of classes in a meta-learning framework.~The proposed method has mainly three key features compared to prior works. First,  it removes any dependence on heuristic formulations thanks to its ability to learn any suitable difficulty formulation for a given dataset. Second, it estimates relative difficulty of a class compared to the other classes whereas prior works use only absolute difficulty of a class in question. Third, it employs a new driver loss function that helps to drive Difficulty-Net learning in a reasonable direction. We verified the effectiveness of the proposed method by conducting extensive experiments on multiple datasets. Further analysis also demonstrated the usefulness of relative difficulty and the newly proposed driver loss function.


{\small
\bibliographystyle{ieee_fullname}
\bibliography{egbib}
}

\end{document}


\title{Difficulty-Net: Learning to Predict Difficulty for Long-Tailed Recognition \\ Supplementary Material}


\maketitle
\thispagestyle{empty}

\setcounter{figure}{4}
\setcounter{table}{4}
\section{More implementation details}
\subsection{More details on datasets}
Table \ref{tab:num_samples} shows the number of samples in training, validation, and test splits in each dataset.
As shown in the table, meta-learning (ML) based methods including ours and \cite{jamal,meta-weight-net} reuse validation images for constructing $S^{meta}$, a dataset to be used for meta learning.
Note that (1) our proposed method was compared with other ML based methods in exactly the same conditions, (2) we re-ran the experiments using public codes of previous methods and therefore all the methods compared in this paper are evaluated using exactly the same split as shown above, and (3) all the ML based methods including ours do not use any extra data, and therefore they do not receive any unfair benefit compared to other methods by having $S^{meta}$.
All the hyper-parameters were tuned using the validation sets.

\begin{table}
  \begin{center}
    {\small{
\begin{tabular}{llllr}
  \toprule
            & Imb. & Train & Val / Meta & Test \\ \midrule
        CIFAR-LT    & 10--200& 49--2 / 490 & 10 & 100  \\
        ImageNet-LT  & 256  & 5 / 1280 & 10 & 50  \\
        Places-LT    & 996 & 5 / 4980 & 10 & 100  \\ 
    \bottomrule     
    \end{tabular}
    }}
    \end{center}
    \caption{The imbalance ratio (Imb.) and the number of samples per class in each datasets. Since the training splits are imbalanced, we show the number of samples in the least and most frequent classes. Note that exactly the same set of images are used for both the validation sets and meta sets. 
    This indicates that meta-learning based methods including ours do not exploit any extra data.}
    \label{tab:num_samples}
\end{table}

\subsection{Hyperparameter settings}
For CIFAR100-LT, following \cite{eqlloss,PaCo}, we use AutoAugment \cite{AutoAugment} and Cutout \cite{Cutout}. Following \cite{eqlloss}, we train ResNet-32 \cite{ResNet} for 12.8K steps with a batch size of 128 and an initial learning rate of 0.1. The learning rate is linearly warmed up to 0.2 over the first 400 steps. It is also decayed by 0.1 after 6.4K and 9.6K steps. For the classifier learning stage in decoupled learning methods, we fix the feature extractor and re-train the classifier for 50 steps using class-balanced sampling following \cite{decoupling}. The learning rate used is 0.1 and is decayed by 0.1 after 30 and 40 epochs. For both the stages, we use a weight decay of $1e-4$. CIFAR100-LT experiments are done on a single NVIDIA Tesla V100 GPU.

For ImageNet-LT, we follow \cite{mislas} and train the models for 180 steps with an initial learning rate of 0.05. The batch size used is 128. We use cosine learning rate decay and weight decay of $5e-4$. In decoupled training, for the second stage we only re-train the classifier for 10 steps using batch size 128 and cosine decayed learning rate with an initial value of 0.05. The models are trained on four NVIDIA Tesla V100 GPUs.

For Places-LT, following \cite{mislas,oltr} we load a ResNet-152 pretrained on ImageNet and then finetune it for 30 steps using an initial learning rate of 0.01 and weight decay of $5e-4$. The learning rate is decayed by 0.1 after 10 and 20 steps. The batch size used is 128. For the classifier learning stage, we retrain the classifier for 20 steps with a batch size of 256 and initial learning rate of 0.1, which is cosine decayed. The training is done on four NVIDIA Tesla V100 GPUs. 

For all the datasets, we use SGD optimizer with momentum 0.9. For Difficulty-Net learning, we use ADAM optimizer with a learning rate of 0.001 and a weight decay $1e-4$. All implementations are done on PyTorch.

\subsection{Designing the Difficulty-Net}
\begin{figure}
    \begin{center}
    \includegraphics[width=\linewidth]{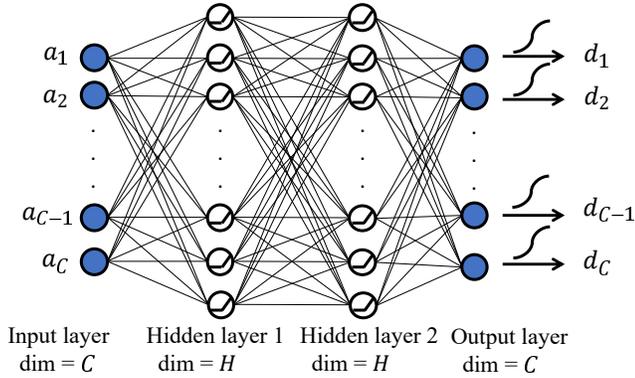}
    \caption{Illustration of our Difficulty-Net}
    \label{fig:difficulty-net_illus}
    \end{center}
\end{figure}

As stated in Sec.~\ref{difficultynet design}, our Difficulty-Net is a MLP with 2 hidden layers. The illustration of our Difficulty-Net is given in Fig.~\ref{fig:difficulty-net_illus}. 
The output layer dimension changes with the number of classes in the dataset. Here we provide a simple way to select the hidden layer dimensions $H$.  To come up with the method, we compare the end-to-end training performance using different values for $H$ on 2 different datasets. The results are given in Table~\ref{tab:effect_of_H}.

\begin{table}
  \begin{center}
    {\small{
\begin{tabular}{lll}
   \toprule
         \multirow{2}{*}{$H$} & CIFAR100-LT & ImageNet-LT \\
          & ($C = 100$) & ($C = 1000$) \\
         \midrule
         128 & \underline{47.96} & 39.4 \\
         256 & \textbf{48.06} & 40.4 \\
         512 & 47.81 & \underline{41.2} \\
         1024 & 47.34 & \textbf{41.4} \\
         2048 & 46.92 & 40.8 \\
    \bottomrule     
    \end{tabular}
    }}
    \end{center}
    \caption{Effect of $H$ on e2e training of ResNet-32 and ResNet-10 on CIFAR100-LT (imbalance=100) and ImageNet-LT respectively.}
    \label{tab:effect_of_H}
\end{table}
    
     

\setlength{\tabcolsep}{1.4pt}
We find that the best working $H$ is different for different datasets. Therefore, based on the results, we decide to select $H=2^n$ such that $2^{n-1} \leq C < 2^{n}$, where $C$ is the number of classes and $n$ is a positive integer. \\
The value of $C$ and $H$ for the three different datasets that we used are given in Table~\ref{tab:H_values}.
\begin{table}
  \begin{center}
  {\small{
  \begin{tabular}{lll}
    \toprule
    Dataset & $C$ & H \\
    \midrule
  CIFAR100-LT & 100 & 128 \\
    ImageNet-LT & 1000 & 1024 \\
    Places-LT & 365 & 512  \\
     
    \bottomrule
  \end{tabular}
  }}
  \end{center}
   \caption{$C$ and $H$ for datasets used in our experiments. }
  \label{tab:H_values}
\end{table}

\subsection{Algorithm for meta-learning via Difficulty-Net}
The algorithm for our Difficulty-Net based learning is provided in Algorithm~\ref{difficulty-net learning}. As stated in Sec.~\ref{sec:meta-learning_via_difficulty-net}, our learning method comprises of three main steps (Eq.~\ref{step 1},\ref{step 2} and \ref{step 3}) that are represented by steps 7, 8 and 10 in the algorithm. Note that 
in our algorithm, $S^{meta}$ is reused as validation set $S^{val}$ for calculating accuracies.

\renewcommand{\algorithmicensure}{\textbf{Output:}}
\begin{algorithm}
\caption{Meta-learning using Difficulty-Net}\label{difficulty-net learning}
\begin{algorithmic}[1]
\Require Training set $S^{train}$, Meta dataset $S^{meta}$
\Require Initial learnable parameters $\theta_1$ and $\phi_1$ 
\Require Max iterations $T$, Value of $\lambda$
\Require Learning rates $\alpha$, $\beta$ and batch sizes $b$, $m$
\For{$t = 1 \dots T$}
  \State Compute $A_{C,t}$ using $f(x; \phi_t)$ on $S^{meta}$
  \State Sample mini-batch of size $b$ from $S^{train}$
  \State Sample mini-batch of size $m$ from $S^{meta}$
  \State Compute weights with $A_{C,t}$ and $\theta_t$ using Eq.~\ref{weight computation}
  \State Compute intermediate $\hat{\phi_t}(\theta_t)$ using Eq.~\ref{step 1}
  \State Update $\theta_t$ to $\theta_{t+1}$ using Eq.~\ref{step 2}
  \State Re-compute Eq.~\ref{weight computation} with $A_{C,t}$ and $\theta_{t+1}$
  \State Update $\phi_t$ to $\phi_{t+1}$ using Eq.~\ref{step 3}
  \EndFor
  \Ensure $\phi_{T+1}, \theta_{T+1}$
\end{algorithmic}
\end{algorithm}

\section{More Results}
\subsection{CIFAR100-LT results without using extra augmentations}
For the CIFAR100-LT results reported in Table~\ref{tab:CIFAR-LT results}, we used extra augmentations (AutoAugment \cite{AutoAugment} and Cutout \cite{Cutout}) to ensure same training setups as recent SOTA methods such as PaCo\cite{PaCo}, BALMS \cite{BALMS} and DRO-LT\cite{dro-lt} for fair comparison. As expected, these additional augmentation techniques provide a significant boost in the results. To verify that our proposed method is effective independent of these extra augmentations, we compare the results of our method with other SOTA methods without using the augmentation techniques. The results are reported in Table~\ref{tab:CIFAR-LT results w/o extra augmentation}. With or without extra augmentations, Ours + LAS proves to be very effective. 
\begin{table}
  \begin{center}
    {\small{
\begin{tabular}{lccccc}
\toprule
&\multicolumn{5}{c}{Imbalance}\\
    \midrule
    Method &  200 & 100 & 50 & 20 & 10\\
    \midrule
    \textit{e2e training} \\
    Focal Loss~\cite{focalloss} $\dagger$ & 35.62 & 38.41 & 44.32 & 51.95 & 55.78 \\
    MWN~\cite{meta-weight-net} $\dagger$& 37.91 & 42.09 & 46.74 & 54.37 & 58.46 \\
    Class-Balanced~\cite{classbalancedloss} $\dagger$ &36.23 & 39.60 & 45.32 & 52.99 & 57.99\\
    CB-DA~\cite{jamal} $\dagger$ & 39.31 & 43.35 & 48.53 &55.62 & 59.58\\ 
    LDAM~\cite{ldam-drw} $\dagger$ & -- & 39.60 & -- & -- & 56.91\\
    EQL~\cite{eqlloss} $\dagger$ & 37.34 & 40.54 & 44.70 & 54.12 & 58.32\\
    CDB-CE~\cite{cdb-ce} $\dagger$ & 37.40 & 42.57 & 46.78 & 54.22 & 58.74 \\
    
    PaCo~\cite{PaCo} & 36.96 & 40.92 & 46.97 & 53.66 & 59.59\\
    \hspace{1mm} + Bal.~Softmax~\cite{BALMS} & 39.55 & 44.13 & 48.60 & 55.89 & 60.24 \\
    Ours & 39.94 & 43.82 & 49.00 & 55.70 & 60.25 \\
    \hspace{1mm} + Bal.~Softmax & 41.43 & 45.81 & 51.14 & 56.58 & \underline{61.33} \\
    \midrule
    \textit{decoupled learning}\\
     cRT~\cite{decoupling} & 40.13 & 44.04 & 48.97 & 55.67 & 59.54 \\
     LWS~\cite{decoupling} & 40.70 & 45.05 & 49.70 & 56.22 & 60.00\\
     LAS~\cite{mislas} & 40.76 & 45.32 & 49.96 & 56.66 & 59.96 \\
     BALMS~\cite{BALMS}  & 39.58 & 44.64 & 48.52 & 54.28 & 58.34\\
     MWN + cRT &40.57 & 44.00 &49.47 & 56.05 & 59.64 \\
     MWN + LWS &40.48& 44.52 & 49.10&55.89 & 59.48 \\
     MWN + LAS & 40.94& 44.64 &49.15 &55.91 & 59.24\\
     Ours + cRT & 41.12 & 45.41 & 50.50 & 56.30& 60.86\\ 
     Ours + LWS & \underline{41.67} & \underline{46.04} & \underline{51.27} & \underline{56.66} & 61.30\\ 
     Ours + LAS & \textbf{42.19} &\textbf{46.42} & \textbf{51.60} & \textbf{56.82} & \textbf{61.47}\\
\bottomrule
\end{tabular}
}}
\end{center}
\caption{Top-1 classification accuracy (\%) on CIFAR100-LT without using extra augmentation \ie AutoAugment and Cutout. 
  $\dagger$ denotes copied results from origin paper \cite{cdb-ce,jamal}. The best results are made bold while the second best results are underlined, which applies for the other tables as well.}
  \label{tab:CIFAR-LT results w/o extra augmentation}
\end{table}

\subsection{ImageNet-LT results on many-, med- and few-shot classes}
In Table~\ref{tab:ImageNet-LT results}, we saw that our proposed method helps to achieve the best overall accuracy. Here we study the effectiveness of our method for each of many-, medium- and few-shot classes. 
The comparison results are given in Table~\ref{tab:ImageNet-LT results_classsubset}. 
\begin{table*}[t]
  \begin{center}
  {\small{
  \begin{tabular}{l|lllr|lllr}
    \toprule
    Backbone Network & \multicolumn{4}{|c|}{ResNet-10} & \multicolumn{4}{|c}{ResNet-50} \\
    \midrule
    Method &  Many & Medium & Few & Overall & Many & Medium & Few & Overall \\
    \midrule
    \textit{e2e training} & &&&&&&&\\
    CE & \underline{57.6} & 25.7 & 3.2 & 34.8 & 64.0 & 38.8 & 5.8 & 41.6 \\
    Focal Loss \cite{focalloss} & 36.4 & 29.9 & 16.0 & 30.5 &--&--&--&-- \\
    OLTR \cite{oltr} & 43.2 & 35.1 & 18.5 & 35.6 & -- &-- & --&--  \\
    EQL \cite{eqlloss}&--&--&--& 36.4 &-- &--&--&-- \\
    CDB-CE \cite{cdb-ce}&--&--&--& 38.5 & --&-- & --&--  \\
    Bal.~Softmax \cite{BALMS}& 55.8 & 35.7 & 20.9 & 41.1 & -- & -- & -- & --\\
    PaCo\cite{PaCo}$^*$ &--&--&--&--& \textbf{68.4} & 44.8 & 14.7 & 49.8  \\
    \hspace{1mm} + Bal.~Softmax \cite{BALMS}$^*$ &--&--&--&--& 59.9 & \textbf{52.6} & 36.1 & 53.5  \\
    Ours & \textbf{58.8} & 36.4 & 13.9 & 41.4 &\underline{68.1} &47.2 &21.5 &51.2   \\
    \hspace{1mm} + Bal.~Softmax & 54.6 &41.6 &27.8 & 44.3 &63.6 & 51.4 & 35.8& \underline{53.7} \\
    \midrule
    \textit{decoupled learning} &&&&&&&&\\
     cRT \cite{decoupling} &--&--&--& 41.8 & 58.8 & 44.0 & 26.1 & 47.3  \\
     LWS \cite{decoupling}&--&--&--& 41.4& 57.1 & 45.2 & 29.3 & 47.7  \\
     MiSLAS \cite{mislas} &--&--&--&--& 61.7 & 51.3 & 35.8 &52.7 \\
     BALMS \cite{BALMS}& 50.3 & 39.5 & 25.3 & 41.8 &--&--&--&-- \\
     Ours + cRT & 53.3 & 41.1 & 27.4& 43.6 & 63.2 & 51.8 & 35.2 & 53.5\\ 
     Ours + LWS & 51.6 & \textbf{43.7} & \underline{29.3} & \underline{44.4} & 62.5 & \underline{52.3} & \underline{36.6}& \underline{53.7} \\ 
     Ours + LAS & 51.8 & \underline{43.6} & \textbf{30.2} & \textbf{44.6} &62.9&\textbf{52.6}&\textbf{36.8}&\textbf{54.0}\\
    \bottomrule
  \end{tabular}
  }}
  \end{center}
 \caption{Top-1 accuracy (\%) on many-, medium- and few-shot classes of ImageNet-LT. $^*$ represents results reproduced using author's codes without using RandAugment \cite{randaugment} for fair comparison. Other results are copied from original papers.    } \label{tab:ImageNet-LT results_classsubset}
\end{table*}
We find that in both e2e learning and decoupled learning, Difficulty-Net based weight assignment helps to significantly boost the performance of the few-shot and medium-shot classes.
We believe this result indicates the strong capability of Difficulty-Net based weighting in mitigating biased performance caused by the class imbalance.
Especially, Ours + LAS is the most effective for the few-shot classes, irrespective of the model used.
\subsection{ImageNet-LT Results Using RandAugment}
In Table~\ref{tab:ImageNet-LT results}, we reproduced the results of PaCo \cite{PaCo} without using RandAugment \cite{randaugment} for the sake of fair comparison with all the other methods that do not use RandAugment.
However, the originally reported results in \cite{PaCo} use RandAugment as additional augmentation, which are significantly higher than the reproduced results. This suggests that PaCo is greatly benefited by the use of RandAugment. Therefore, we used RandAugment with our method and compared the results with PaCo in Table~\ref{tab:rand_augment}. 
We only used Ours + LAS for the comparison because Ours + LAS is the best performing decoupled learning method as seen in Table~\ref{tab:CIFAR-LT results},\ref{tab:ImageNet-LT results} and \ref{tab:Places results}. 

From Table~\ref{tab:rand_augment}, we find that using RandAugment benefits our method as well. With or without RandAugment \cite{randaugment}, Ours+LAS outperformed PaCo.
\begin{table}
    \begin{center}
    {\small{
    \begin{tabular}{lcc}
    \toprule
        Method & ResNet-10 & ResNet-50 \\
        \midrule
         PaCo + Bal.~Softmax \cite{PaCo} &--& \underline{57.0} \\
         Ours + LAS & \textbf{46.9} & \textbf{57.4} \\ 
         \bottomrule
    \end{tabular}
    }}
    \end{center}
    \caption{Top-1 accuracy (\%) using RandAugment\cite{randaugment}. Baseline results are copied from the original paper \cite{PaCo}.}
    \label{tab:rand_augment}
\end{table}

\subsection{Comparison on ImageNet-LT with MoE methods}

In Sec.~\ref{sec:main_results}, we did not compare our proposed method directly to mixture of experts (MoE) methods as the latter uses multiple experts while we focus on improving the learning of a single expert. For the fair comparison with MoE methods, we created an ensemble of Difficulty-Net based trained models. For the ensemble creation, we trained two expert models using Ours + LAS decoupled learning. 
The backbone architectures of these two models were kept the same. 
The only difference between these models was that one used a linear classifier and the other used a cosine classifier. 
During inference, we simply took the mean outputs of these two models. The results of this simple ensemble is provided in Table~\ref{tab:moe}.  

As can be seen, although our ensemble comprises of only two expert models, it performs significantly better than 3-experts and 4-experts RIDE \cite{RIDE}. This shows that our proposed Difficulty-Net is effective in learning expert models for MoE methods. However, the current ensemble is heuristic and a detailed research on contribution of Difficulty-Net in MoE is left for the future.

\begin{table}
\begin{center}
  {\small{
    
    \begin{tabular}{lcc@{}}
    \toprule
         Method & ResNet-10 & ResNet-50  \\
         \midrule
         LFME \cite{LFME} & \underline{38.8} & --\\
          RIDE (2 experts) \cite{RIDE} &--& 54.4 \\
          RIDE (3 experts) \cite{RIDE}&--& 54.9 \\
          RIDE (4 experts) \cite{RIDE}&--& \underline{55.4} \\
          Ours (2 experts) &\textbf{47.5} &\textbf{56.2} \\
          \bottomrule
    \end{tabular}
    }}
    \end{center}
    \caption{Comparison with mixture of expert methods. Baseline results are copied from the original papers \cite{LFME,RIDE}.}
    \label{tab:moe}
\end{table}

\subsection{More results on sample-level v/s class-level difficulty}
As empirically verified in Table~\ref{tab:ablation}, class-level difficulty is more effective than sample-level difficulty in our Difficulty-Net. We believe that this happens because as stated in \cite{cdb-ce} and Sec.~\ref{sec:ablation}, the absolute number of hard samples in head classes is significantly higher than that in tail classes due to the inherent long-tail characteristic of the dataset. Using sample-level difficulty gives high weights to all the hard samples irrespective of their classes, resulting in more weights for the head classes and therefore getting the model biased to the head classes. 

We verified this by conducting a simple experiment on CIFAR100-LT. For 2 classes A and B with 376 and 46 training samples respectively, the absolute number of hard samples given high weights by sample-level method
was higher for A(50) than B(13). Although higher proportion of samples in B($\approx28\%$) received high weights compared to A ($\approx13\%$),
%
A got more weights compared to B due to its higher absolute number of hard samples.
As a result, the accuracy for A is improved from 46\% to 62\% and that for B is decreased from 31\% to 21\%, hence boosting the 
bias. In such case, using class-level difficulty gives high weights to all samples of B, resulting in more weights for B. As a result, the accuracy on B was improved from 31\% to 40\%, while that on A was almost maintained (46\% to 44\%). 

The effectiveness of class-level difficulty in Difficulty-Net for overcoming model bias is further verified in Table~\ref{tab:sample-level_v/s_class_level}. Using sample-level difficulty causes the model to get biased towards the many-shot classes while class-level difficulty is particularly useful for improving performance on the med-shot and few-shot classes.
\begin{table}
    \begin{center}
    {\small{
    \begin{tabular}{l|llll|llll}
    \toprule
    Imbalance & \multicolumn{4}{c|}{100} & \multicolumn{4}{c}{10}\\
    \midrule
     Difficulty    &  Many & Med & Few & All & Many & Med & Few & All\\
     \midrule
     Sample-level  &  \textbf{67.00} & 47.46 & 18.99 & 45.76 & \textbf{74.50} & 61.01 & 48.02 & 62.51\\
     Class-level (ours) & 64.47 & \textbf{51.21} & \textbf{24.92}&  \textbf{47.96} & 70.48 & \textbf{64.40} & \textbf{53.36} & \textbf{63.52}\\
     \bottomrule
    \end{tabular}
    }}
    \end{center}
    \caption{Comparison of sample-level difficulty and class-level difficulty on CIFAR100-LT.}
    \label{tab:sample-level_v/s_class_level}
\end{table}

\makeatletter\@input{xx.tex}\makeatother